\documentclass[lettersize,journal]{IEEEtran}
\usepackage{amsmath,amsfonts}
\usepackage{algorithmic}
\usepackage{algorithm}
\usepackage{array}
\usepackage[caption=false,font=normalsize,labelfont=sf,textfont=sf]{subfig}
\usepackage{textcomp}
\usepackage{stfloats}
\usepackage{url}
\usepackage{verbatim}
\usepackage{graphicx}
\usepackage{multirow}
\usepackage{cite}
\usepackage{color}
\usepackage{hyperref}

\DeclareRobustCommand\onedot{\futurelet\@let@token\@onedot}
\def\eg{\emph{e.g., }}  

\def\ie{\emph{i.e., }}    
   
\def\etc{\emph{etc. }}  
            
\def\etal{\emph{et al.} }
\makeatother

\begin{document}

\title{Blind Face Restoration for Under-Display Camera via Dictionary Guided Transformer}

\author{Jingfan Tan,
        Xiaoxu Chen,
        Tao Wang,
        Kaihao Zhang,
        Wenhan Luo,
        Xiaocun Cao\\
% <-this % stops a space
\IEEEcompsocitemizethanks{\IEEEcompsocthanksitem J. Tan, X. Chen, W. Luo and X. Cao are with Shenzhen Campus of Sun Yat-sen University, Shenzhen, China, 
(e-mail: \{tjfky2001, chenxiaoxu89, whluo.china\}@gmail.com, caoxiaochun@mail.sysu.edu.cn).
\IEEEcompsocthanksitem T. Wang is with the State Key Laboratory for Novel Software Technology, Nanjing University, Nanjing, China, (e-mail: taowangzj@gmail.com).
\IEEEcompsocthanksitem K. Zhang is with the College of Engineering and Computer Science, Australian National University, Canberra, Australia, (e-mail: super.khzhang@gmail.com).

}

}

% The paper headers
\markboth{Journal of \LaTeX\ Class Files,~Vol.~14, No.~8, August~2023}%
{Shell \MakeLowercase{\textit{et al.}}: A Sample Article Using IEEEtran.cls for IEEE Journals}

\maketitle
\begin{abstract}
By hiding the front-facing camera below the display panel, Under-Display Camera (UDC) provides users with a full-screen experience. However, due to the characteristics of the display, images taken by UDC suffer from significant quality degradation. Methods have been proposed to tackle UDC image restoration and advances have been achieved. There are still no specialized methods and datasets for restoring UDC face images, which may be the most common problem in the UDC scene. To this end, considering color filtering, brightness attenuation, and diffraction in the imaging process of UDC, we propose a two-stage network UDC Degradation Model Network named UDC-DMNet to synthesize UDC images by modeling the processes of UDC imaging. Then we use UDC-DMNet and high-quality face images from FFHQ and CelebA-Test to create UDC face training datasets FFHQ-P/T and testing datasets CelebA-Test-P/T for UDC face restoration. We propose a novel dictionary-guided transformer network named DGFormer. Introducing the facial component dictionary and the characteristics of the UDC image in the restoration makes DGFormer capable of addressing blind face restoration in UDC scenarios. Experiments show that our DGFormer and UDC-DMNet achieve state-of-the-art performance. 
\end{abstract}


\section{Introduction}
In recent years, the technique of Under-Display Camera (UDC) has been widely used in smart devices, which can provide a better experience for users. For example, the smartphone using UDC can achieve true full screen without punch holes or notches. Laptops or tablets that place cameras under the center of the displays can enable a more natural gaze to focus when people use the camera. However, placing the camera below the display (\eg OLED screen) can seriously degrade the quality of the images, \eg blur \cite{zhang_mc_blur}, noise \cite{jiang_denoise_2023}, low light \cite{wang2023ultra}, \etc  Enhancing imaging quality can be achieved through the redesign of the display panel's physical structure, including optimizations in the spatial arrangement of the opening. However, the development costs associated with the refining of OLED display panels can be prohibitively high. An alternative and cost-effective approach involves harnessing deep learning techniques for the restoration of images captured via Under-Display Cameras (UDC). Consequently, image restoration beneath UDC has emerged as a prominent focal point within the realm of computer vision.

%%%%% existing udc restoration images  
A large number of methods~\cite{MPGNet, DISCNet, ResNet, PDCRN, BNUDC,kwon2021controllable, UDCIR, UDCUNet} have been proposed for image restoration under UDC. For example, DISCNet~\cite{DISCNet} puts the prior knowledge of an accurately measured PSF in the network design to help restore UDC images. Koh \etal \cite{BNUDC} first regard the process of recovery of UDC images as low-spatial-frequency restoration and high-spatial-frequency restoration. Then, they propose a two-branch network called BNUDC to deal with the UDC restoration problem. Although these existing UDC restoration methods work well in the natural UDC image restoration problem, these methods do not fully consider UDC face images. Directly applying these general UDC restoration methods to UDC face images, the recovered results may lack the richness of face details. 

%%%% face 
On the other hand, existing blind face restoration methods~\cite{GFPGAN, GPEN, hifacegan, dmdnet, PSFRGAN, Restoreformer, VQFR} trained on face datasets (\eg FFHQ~\cite{ffhq}) cannot restore the UDC face images very well. Because these datasets and methods do not take into account the characteristics of UDC images, where UDC-degraded images generally suffer from diverse and complicated degradation, \eg low light, blur, and noise~\cite{wang2021video, BNUDC, UDCIR, wang2023ultra}. Thus, existing blind face restoration methods are not applicable for restoring UDC face images. Besides, to the best of our knowledge, there is still no benchmark and specialized method for restoring UDC face images, which may be the most common problem in the UDC scene. 

\begin{figure*}[t] 
\centering
\includegraphics[width=0.92\textwidth]{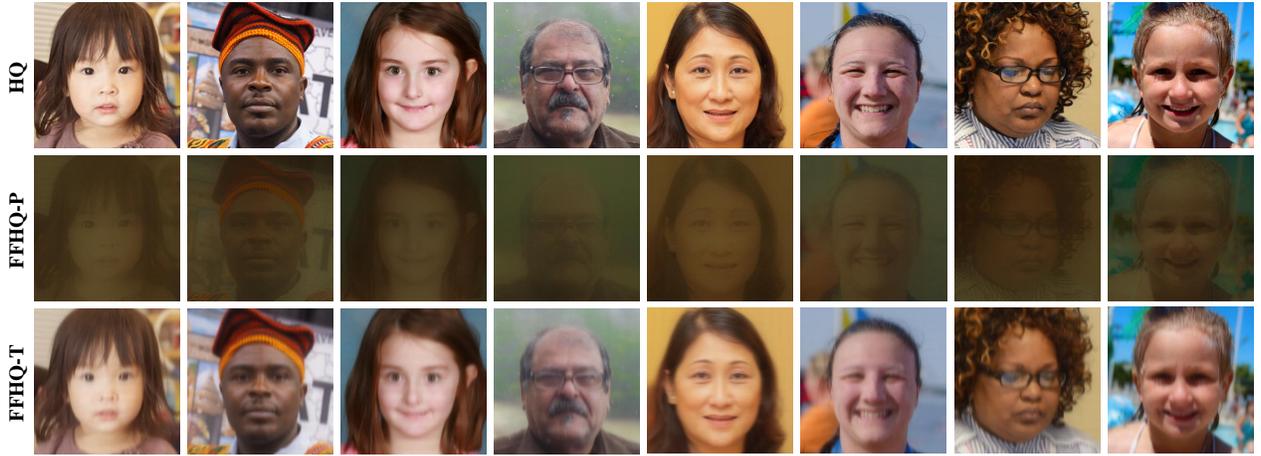}
\caption{Exemplar UDC face images from our built datasets FFHQ-P and FFHQ-T. As observed, the synthesized face images suffer from typical degradation artifacts in the UDC scenery such as low light, blur, and noise.}
\label{fig0}
%\vspace{-4mm}
\end{figure*}

To investigate the UDC face restoration problem, we first propose a synthesis method for synthesizing UDC images with reference to the UDC imaging process. Based on our synthesis method, we build the first UDC face image datasets using the ordinary face dataset FFHQ~\cite{ffhq} and CelebA-Test dataset~\cite{celeba}. Finally, we propose a face restoration method specifically for the UDC scene. Specifically, for UDC degradation synthesis, inspired by previous works~\cite{DAGF,MPGNet}, we propose an end-to-end image synthesis method named UDC Degradation Modeling Network (UDC-DMNet) to simulate the UDC imaging degradation process from existing paired real-world UDC dataset~\cite{ECCVchallenge}. In particular, UDC-DMNet is a two-stage network, and its two stages are designed for modeling the color filtering and diffraction degradation processes of UDC images, respectively. In addition, the network also uses cross-stage affine transformation and feature fusion to connect the two stages. 

To construct UDC face datasets, we first train our UDC-DMNet on two UDC image datasets T-OLED and P-OLED~\cite{ECCVchallenge}, respectively. Then we use $70,000$ high-quality face images from FFHQ~\cite{ffhq} as inputs to the trained UDC-DMNet to synthesize corresponding low-quality face images from the training dataset named FFHQ-P and FFHQ-T for UDC face restoration. In the end, we use $3,000$ high-quality face images from CelebA-Test~\cite{celeba} dataset to synthesize corresponding low-quality face images of the testing dataset, CelebA-Test-P, and CelebA-Test-T. Exemplar images of our built dataset are shown in Fig.~\ref{fig0}.

Based on our built UDC face datasets, we propose a Dictionary-Guide Transformer called DGFormer for UDC face restoration. DGFormer uses the facial component dictionary as the face prior and takes into account the characteristics of UDC images to achieve effective UDC face restoration. To be specific, our DGFormer consists of four modules: Shallow Feature Extractor (SFE), UDC Restoration Module (UDCRM), Dictionary-Guided Restoration Module (DGRM), and Image Reconstruction Module (IRM). SFE extracts image features and UDCRM achieves coarse-grained UDC degradation removal by several transformer blocks designed considering UDC characteristics. DGRM adopts the proposed dictionary-guided transformer blocks containing facial dictionaries, which can provide diverse facial component textures for fine-grained restoration. IRM reconstructs the fine-grained feature to a clear image. The comprehensive experiments show that our UDC-DMNet and our DGFormer achieve state-of-the-art performance. We hope that DGFormer and UDC-DMNet can benefit the community.

Overall, our contributions can be summarized as follows:
\begin{itemize}
    \item We propose UDC-DMNet to simulate the UDC imaging degradation process from the existing paired real-world UDC dataset. With UDC-DMNet, we build the first UDC face restoration datasets, FFHQ-P/T for training and CelebA-Test-P/T for evaluation of UDC face. To the best of our knowledge, we are the first to build UDC face datasets.
    \item We propose a method called DGFormer for UDC face image restoration. DGFormer comprehensively considers the face prior information and the characteristics of the UDC image, achieving effective UDC face restoration. 
    \item According to extensive experimental studies, the UDC image synthesis method UDC-DMNet and the UDC face restoration method DGFormer are state-of-the-art compared with existing methods.
\end{itemize}

The subsequent sections of this paper are structured as follows: In Section \ref{sec:related_work}, we review pertinent research encompassing UDC imaging, UDC image restoration, and blind face restoration. Section \ref{sec:methodology} illustrates our UDC-DMNet and DGFormer models. Section \ref{sec:experiment} provides a comprehensive account of the experimental results and their corresponding analyses. Ultimately, our conclusion and future work are encapsulated in Section \ref{sec:conclusion}.

\section{Related Work}
\label{sec:related_work}
In this section, we present a brief literature review of UDC imaging, UDC image restoration, and blind face restoration.
\subsection{UDC Imaging and Degradation Model}
Zhou \etal \cite{UDCIR} first analyze the UDC optical system and then establish the physical model of the degradation of images captured by UDC. Specifically, in the UDC imaging system, before being captured by the sensor, the light emitted from a point light source is modulated by the OLED and camera lens. Due to the diffraction effects, light transmission rate, camera noise, and other problems caused by OLED and camera lenses, UDC images always tend to be blurry, low-light, and noisy. The degradation processing of UDC imaging is typically modeled by
\begin{equation}
    y=(\alpha x)*k + n,
\label{eq1}
\end{equation}
where $x$ is the clean real world image, and $y$ is the degraded UDC image. $\alpha$ represents the intensity scaling factor which simultaneously relates to the physical light transmission rate of the display and the changing ratio of the average pixel values. $k$ is the Point Spread Function (PSF) which depends on the type of display. $n$ is additive noise, and $*$ denotes the convolution operator. Based on the above model, several works~\cite{DISCNet, UDCUNet, kwon2021controllable} make various changes considering different aspects of UDC. For example, considering proper dynamic range for the scenes and camera sensor, Feng \etal \cite{DISCNet} turn the input image into an HDR image. Considering that the PSF is not valid for the corners of the image, where the light is obliquely incident on the panel, Kown \etal \cite{kwon2021controllable} use PSF from different angles. Besides, Koh \etal \cite{BNUDC} consider the different transmission rates of wavelengths by the thin-film layers of an OLED and further simplify the degradation process into two stages. It can be formulated as
\begin{equation}
    y=\Psi(\Phi(x)),
\label{eq2}
\end{equation}
where $\Phi$ represents color filtering and spatially variant attenuation caused by thin-film layers of the OLEDs and $\Psi$ is diffraction which can be represented by a PSF caused by the pixel definition layer of the OLEDs.

\begin{figure*}[t]
\centering
\includegraphics[width=0.95\textwidth]{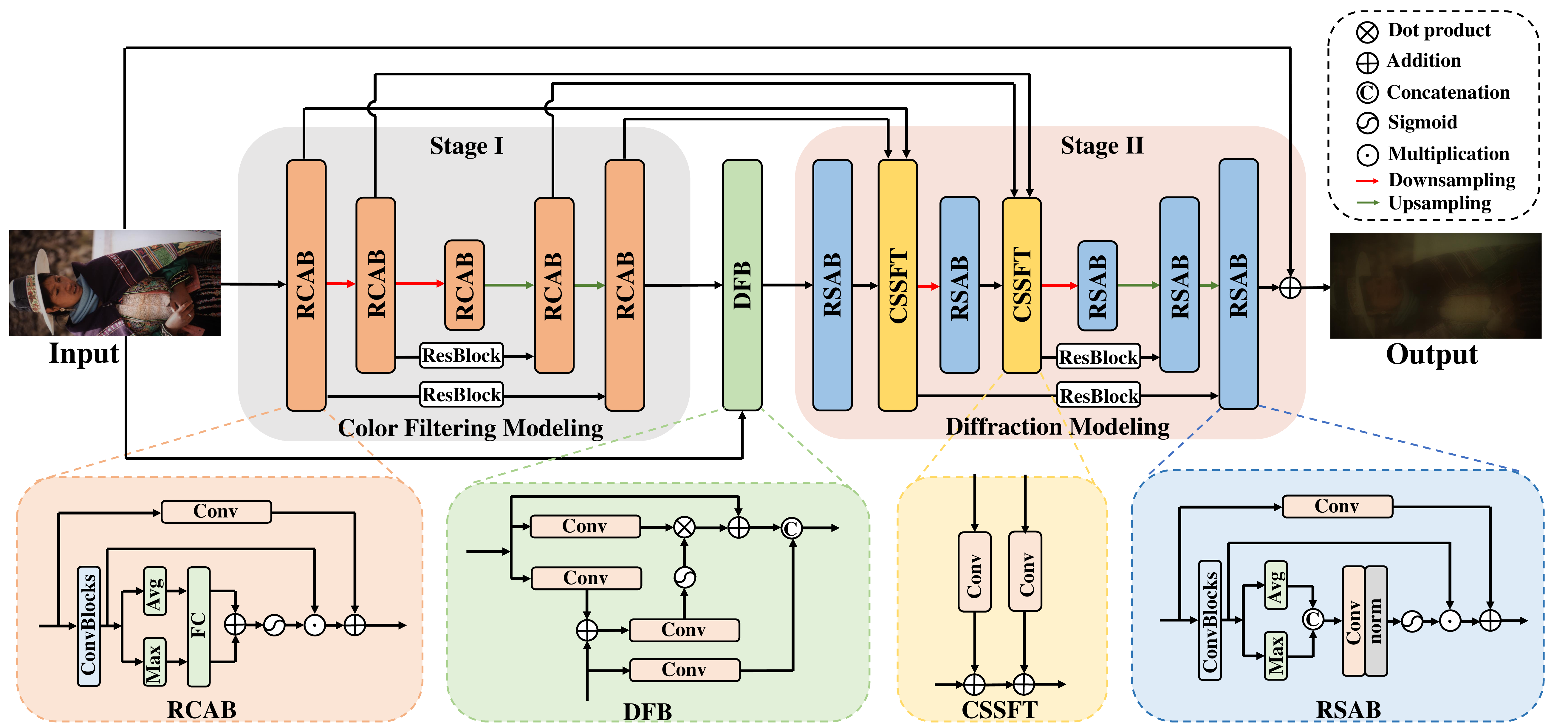}
%\vspace{-2mm}
\caption{The architecture of the proposed UDC-DMNet framework, which consists of two encoder-decoders to model the degradation of UDC images. The basic block of Stage-\uppercase\expandafter{\romannumeral1} is the Residual Channel Attention Block (RCAB). The basic block of Stage-\uppercase\expandafter{\romannumeral2} is the Residual Spatial Attention Block (RSAB). Between the two stages, we adopt a Degradation Fusion Block (DFB). The Cross Stage Spatial Feature Transform (CSSFT) is also used for connecting two stages. }
\label{fig1}
% \vspace{-2mm}
\end{figure*}

\subsection{UDC Image Restoration}
In the work of UDC image restoration, Zhou \etal \cite{UDCIR} propose a UDC dataset with paired low-quality and high-quality images which are captured with Monitor-Camera Imaging System (MCIS) devised by the authors. There are some works in the ECCV challenge~\cite{ECCVchallenge} trying to solve the UDC image restoration problem. Feng \etal \cite{DISCNet} construct a new UDC image dataset that uses HDR images as input and measures the PSF of the UDC system to generate UDC images. They also propose DISCNet to address the UDC image restoration problem. The MIPI challenge \cite{mipi} aims to tackle the problem of UDC image restoration and attracts more and more attention.
Besides, by achieving a deep learning approach that performs a RAW-to-RAW image restoration, Qi \etal \cite{qi2021isp} propose an image-restoration pipeline that is ISP-agnostic and successfully accomplishes superior quantitative performance. Kwon \etal \cite{kwon2021controllable} tackle this problem as denoising and deblur utilizing pixel-wise UDC-specific kernel representation and a noise estimator, achieving higher perceptual quality. Koh \etal \cite{BNUDC} introduce a deep neural network with two branches to reverse each type of degradation of UDC. Their network outperforms existing methods on all three datasets of UDC images. Considering the deployment of the mobile terminal, Li \etal \cite{LightUDC} propose a lightweight method and further distill the proposed model.

Although the above methods have been used to deal with the natural UDC image restoration problem, there is still no specialized method or dataset for the restoration of face images in the UDC scenery, which may be the most common problem in the UDC scene. Thus, we propose a method for synthesizing UDC images with reference to the UDC imaging process. With this synthesis method, we build UDC facial image datasets using the ordinary face dataset FFHQ~\cite{ffhq} and CelebA-Test~\cite{celeba} and propose a face restoration method specifically for the UDC scene.

\subsection{Blind Face Restoration}
Due to the wide application scenarios of face images, face image restoration has received a lot of attention in all aspects. \eg face super resolution~\cite{FSR1, FSR2, FSR3}, blind face restoration~\cite{GFPGAN, DFDNet, zhang1}, face inpainting~\cite{inpaint1, inpaint2, inpaint3}, \emph{etc}. Among them, blind face restoration (BFR) aims to restore high-quality face images from low-quality ones without the knowledge of degradation types or parameters~\cite{WangTaosurvey}. In the literature, BFR methods can be approximately divided into two categories: prior-based restoration methods and non-prior-based restoration methods. Some non-prior-based methods~\cite{hifacegan} perform BFR by learning the mapping function between low-quality and high-quality facial images. Because of the high degree of face structures, most blind face restoration methods use face priors to recover facial images with clearer facial structures, such as generative prior~\cite{Pulse, DEARGAN, GPEN, SGPN}, reference prior~\cite{DFDNet, VQFR, Restoreformer,asff}, and geometric prior~\cite{Fsrnet,bulat2018super, yu2018face, zhang2}.

Generative prior used in BFR methods mainly include using GAN inversion~\cite{Pulse} and pre-training facial GAN models~\cite{GFPGAN, GPEN, DEARGAN} such as StyleGAN~\cite{style} to provide richer and more diverse facial information. However, these methods based on generative prior do not consider identity information during the training and the methods are limited to the latent space of generators. Although these methods show good performance in terms of image generation metrics and detail texture, their restored images lack fidelity. Several reference prior based methods~\cite{DFDNet, Restoreformer, asff, VQFR} guide the face restoration process by using the facial structure or facial component dictionary obtained from additional high-quality face images as a reference prior. Li \etal \cite{DFDNet} propose a deep face dictionary network (DFDNet) for face restoration, which uses a dictionary of facial components extracted from high-quality images as a reference prior. Then, they select the most similar component features from the component dictionary to transfer the details to low-quality face images for face restoration. While VQFR~\cite{VQFR} and RestoreFormer~\cite{Restoreformer} using VQGAN~\cite{vqgan} pretrain a high-quality codebook on entire faces, acquiring rich expressiveness. The unique geometric shape and spatial distribution information of faces in the images are used to help the model gradually recover high-quality face images in geometric prior-based methods, which mainly include facial landmarks~\cite{Fsrnet,kim2019progressive}, facial heatmaps~\cite{yu2018face}, and facial parsing maps~\cite{PSFRGAN}. In these methods, geometric priors usually need to be extracted from degraded face images, but the degraded face images cannot obtain the prior information accurately. In addition, geometric priority cannot provide rich details for face restoration.

\section{Methodology}
\label{sec:methodology}
In this section, We first introduce our method for synthesizing face UDC images. Based on the proposed synthetic method, we build the first UDC face datasets (\ie FFHQ-P/T for training, and CelebA-Test-P/T for testing) by using FFHQ and CelebA-Test datasets. 
Moreover, with these synthetic UDC face datasets, we further propose a dictionary guide transformer restoration network called DGFormer for blind face restoration in UDC.

\subsection{UDC Degradation Synthesis and Datasets}

\subsubsection{Degradation Synthesis}
Existing methods usually use the UDC degradation model in Eq.~\eqref{eq1} to synthesize UDC-degraded images. However, there exists a big gap between this approximate degradation model and the real UDC degradation. For example, the parameters in degradation (intensity scale factor $\alpha$ and PSF $k$) are different in UDC configurations, which requires extensive human effort to obtain these parameters in advance~\cite{UDCIR, DISCNet,kwon2021controllable}. 
Thus, we design an end-to-end UDC degradation modeling network called UDC-DMNet to directly learn the UDC degradation process with the existing real-world paired data. 

As shown in Fig.~\ref{fig1}, our UDC-DMNet models the UDC degradation process as two stages~\cite{BNUDC} \ie \textbf{color filtering and brightness attenuation modeling}, and \textbf{diffraction modeling}. 
Correspondingly, our UDC-DMNet is a two-stage network, where the first stage is used for color filtering and brightness attenuation modeling, and the second stage is designed for diffraction modeling.

\textbf{Color Filtering and Brightness Attenuation Modeling}. When the light is normally incident on the OLED display, part of the light will be absorbed by the display, because the thin-film layers of an OLED have different transmission rates for different wavelengths. Thus, a clean image first exhibits color filtering and brightness attenuation in the UDC imaging process and the degradation is highly related to the channels of the image. To model this degradation process, as shown in Fig.~\ref{fig1}, we first use one $3 \times 3$ convolutional layer in the first stage of our UDC-DMNet to extract the initial features. Then, those features are fed into an encoder-decoder architecture with four downsampling and upsampling operations to derive multi-scale information. In each scale, we use a Residual Channel Attention Block (RCAB)~\cite{rcab,cbam} to adaptively learn the process of brightness attenuation. In particular, RCAB consists of three convolution layers and one channel attention layer. With the help of RCAB, the first stage of our UDC-DMNet can fully simulate the brightness and color of the UDC image.

\begin{figure*}[t]
\centering
\includegraphics[width=0.95\textwidth]{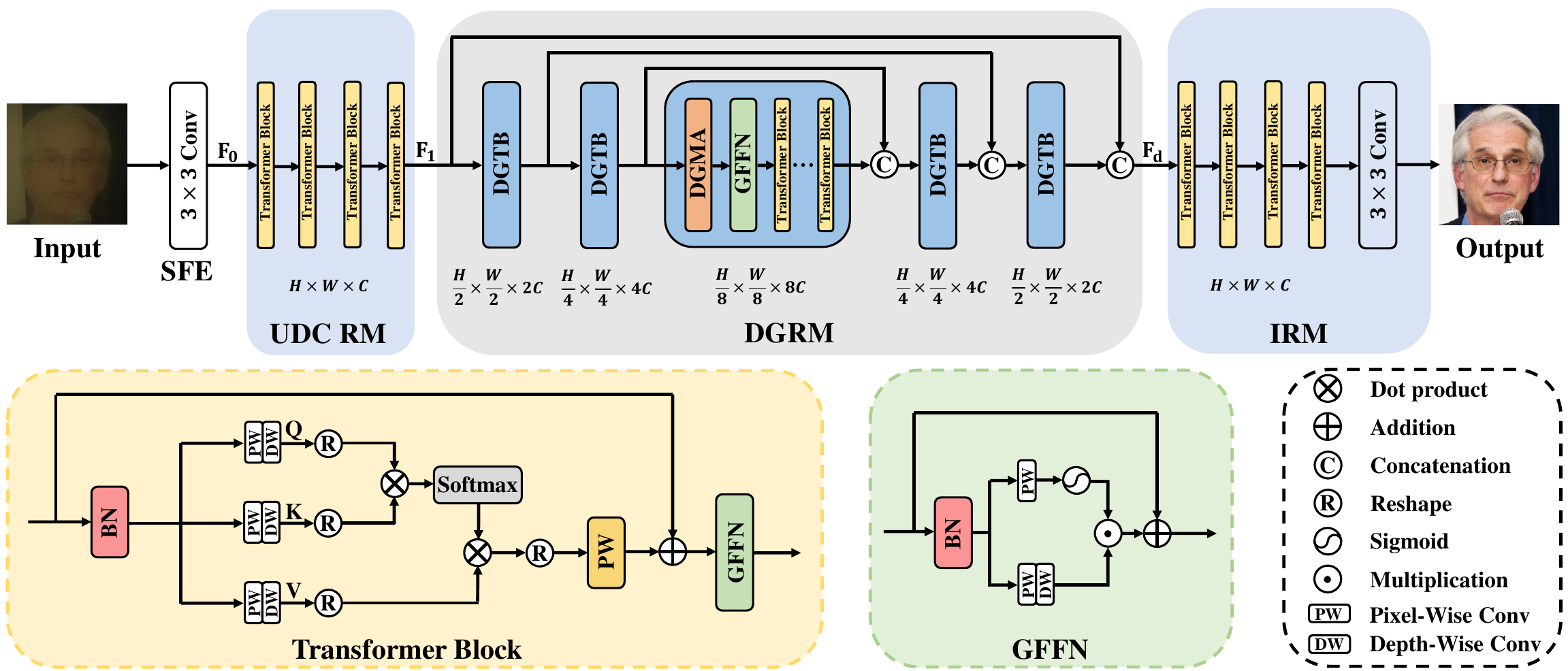}
    %\vspace{-2mm}
\caption{The architecture of the proposed  DGFormer, which consists of a Shallow Feature Extractor (SFE), a UDC Restoration Module (UDCRM), a Dictionary-Guided Restoration Module (DGRM), and an Image Reconstruction Module (IRM). The basic block of DGRM is the proposed Dictionary-Guided Transformer Block (DGTB), which contains one Dictionary-Guided Multi-head Attention (DGMA), one Gate Feed Forward Network (GFFN), and several transformer blocks.}
\label{fig2}
%\vspace{-4mm}
\end{figure*}
\textbf{Diffraction Modeling}. In the second stage of UDC imaging, the light will be incident on the pixel definition layer (PDL). PDL that has a periodic window pattern acts as a planar diffraction grating~\cite{pdl,pdl2}. And images formed under PDL will be blurry. Previous works~\cite{UDCIR,DISCNet,kwon2021controllable} usually estimate the PSF to simulate this process. However, PSF is highly complicated, and the measured PSF is usually not universal. Thus, in this work, we also choose encoder-decoder architecture to model this diffraction process. To better cover the PSF of various scales, we use a convolution with the stride of 2 to downsample and use a transposed convolution to upsample the feature maps. Specifically, in each scale, we adopt a residual block with spatial attention \cite{cbam} named Residual Spatial Attention Block (RSAB) as the basic layer to build the second stage of our UDC-DMNet.
With the help of RSAB, our UDC-DMNet can well simulate the blur caused by spatial changes in the second stage. 

Moreover, considering that the degradation of the first stage may affect the diffraction of the second stage and a long series of convolution operations are likely to cause blurring of edge features. Thus, as shown in Fig.~\ref{fig1}, we propose a Cross-Stage Spatial Feature Transform (CSSFT) to connect two stages of our UDC-DMNet. CSSFT can help simplify the information flow among stages to address the above problem~\cite{MPRNet,wang2018recovering}. Specifically, CSSFT can be formulated as follows,
\begin{equation}
    {\alpha}_{e},{\alpha}_{d} = Conv1(F_e), Conv2(F_d), 
\end{equation}
where $Conv1(\cdot)$ and $Conv2(\cdot)$ are both $1\times 1$ convolution. $F_e$ and $F_d$ are the same scale features from the encoder and decoder in the first stage, respectively. ${\alpha}_{e}$,\ ${\alpha}_{d}$ are the parameters of the spatial feature transform, which can be defined as
\begin{equation}
    F_{out} = F_{in} + {\alpha}_{e} + {\alpha}_{d},
\end{equation}
where $F_{in}$ is the output of RSAB, and $F_{out}$ represents the output of the spatial feature transform.

In addition, as shown in Fig.~\ref{fig1}, we introduce a Degradation Fusion Block (DFB) between the first and second stages of our UDC-DMNet to fuse color-filtering degraded features with the input image, which makes the second stage of UDC-DMNet better focus on diffraction modeling.

\subsubsection{Loss Function}
We first select the Peak Signal-to-Noise Ratio (PSNR) loss as the main supervision~\cite{hinet}. However, the PSNR loss function merely measures the pixel-wise distance and cannot properly describe the similarity of the degradation of two images. The generated degraded image will be over-smooth. Therefore, we further select perceptual loss~\cite{perceptual} and adversarial loss~\cite{GAN} to learn the degradation patterns in training. We adopt three losses to train UDC-DMNet: PSNR loss $\mathcal{L}_{\text{psnr}}$, perceptual loss $\mathcal{L}_{\text{per}}$ and adversarial loss $\mathcal{L}_{\text{adv}}$.
The PSNR loss is defined as:
\begin{equation}
\mathcal{L}_{\text{psnr}}=PSNR(\hat{\mathbf{X}}, \mathbf{I}),
\end{equation}
where $\hat{\mathbf{X}}$ and $\mathbf{I}$ refer to the generated degraded image and ground-truth degraded image respectively. Then the perceptual loss is used to help our model produce visually pleasing results. We adopt a pre-trained VGG19~\cite{VGG} to extract the perceptual features from the $Conv5\_4$ layer of VGG19, and then use the $L_{1}$ loss function to calculate the difference in the feature space between the generated degraded image and their corresponding ground-truth degraded image. Specifically, the perceptual loss is as follows:
\begin{equation}
	\begin{aligned}
	\mathcal{L}_{per} =  \| VGG(\hat{\mathbf{X}})-VGG (\mathbf{I}) \|_{1}.
	\end{aligned}
\end{equation}
We use adversarial loss on the output of the generator of UDC-DMNet to distinguish real and fake degraded images. This is helpful to make the UDC-DMNet model the degradation process better and narrow the gap between the generated distribution and the real-world distribution. Specifically, we adopt the discriminator $\mathcal{D}$ of PatchGAN~\cite{cyclegan} for adversarial training:
\begin{equation}
	\begin{aligned}
	\mathcal{L}_{adv} =  [log\mathcal{D}(\mathbf{I})+log(1-\mathcal{D}(\hat{\mathbf{X}}))].
	\end{aligned}
\end{equation}
The final loss function $\mathcal{L}$ to train our proposed UDC-DMNet is shown as follows:
\begin{equation}
\mathcal{L}=\mathcal{L}_{\text{psnr}}+ {\lambda}_{1} \mathcal{L}_{\text {per}}+ {\lambda}_{2} \mathcal{L}_{\text {adv}},
\end{equation}
where ${\lambda}_{1}$ and ${\lambda}_{2}$ are hyper-parameters used to balance these three losses. In our experiments, they are both set to $1$.

\subsubsection{Building UDC Face Datasets}
We train our UDC-DMNet on the P-OLED and T-OLED datasets~\cite{ECCVchallenge}, respectively. Both datasets have $300$ pairs of degraded UDC images captured using a UDC device in the real world and the corresponding ground truth images.
Each image is of resolution $1024 \times 2048$. 
After the model training, we leverage the FFHQ dataset~\cite{ffhq}, which consists of $70,000$ high-quality face images as inputs to the trained UDC-DMNet to synthesize UDC face datasets, FFHQ-P and FFHQ-T, which will be specifically employed for the training of our proposed UDC face image restoration model illustrated in the following.
For evaluation of UDC face image restoration, we use high-quality face images in CelebA-Test dataset~\cite{celeba} as inputs of the trained UDC-DMNet to synthesize CelebA-Test-P and CelebA-Test-T.

\subsection{Dictionary Guide Transformer}
To better restore facial images in the UDC scenery, we further propose a transformer-based network with dictionary guidance, named DGFormer.

\subsubsection{Overall Architecture}
As shown in Fig.~\ref{fig2}, DGFormer consists of four modules: shallow feature extractor, UDC restoration module, dictionary-guided restoration module, and image reconstruction module. Specifically, give a degraded UDC face image $X \in \mathbb{R}^{H\times W \times 3}$, the shallow feature extractor, consisting of a simple $3 \times 3$ convolution, first extracts features $F_0\in\mathbb{R}^{H\times W\times C}$ from the input image $X$. Then, the feature $F_0$ is fed into the subsequent UDC restoration module to achieve coarse-grained UDC degradation removal. The UDC restoration module is composed of several specially designed transformer blocks. After that, the output feature of the UDC restoration module is forwarded into the dictionary-guided restoration module for fine-grained restoration. The dictionary-guided restoration module is a symmetric encoder-decoder structure, which is mainly built by our proposed dictionary-guided transformer blocks. Finally, the fine-grained feature is reconstructed to a clear image by the image reconstruction module. The UDC restoration module, dictionary-guided restoration module, and image reconstruction module are core modules of our DGFormer, thus we detail them in the following.

\subsubsection{UDC Restoration Module}
The UDC image usually has diverse and complicated degradation (\eg low light, blur, noise \emph{etc.}), thus directly restoring clear images from UDC images is difficult. In addition, the complex degradation in UDC input images causes difficulty in extracting face priors, which influences the performance. Thus, we propose a UDC Restoration Module as the successor of the shallow extractor to achieve coarse-grained UDC degradation removal. Previous work~\cite{MPGNet} claims that the UDC restoration network should consider global and local information in the network design because removing low light and color filtering degradation in UDC images requires global information in each color channel, and local information is more favorable for noise and blur removal. Thus, as shown in Fig.~\ref{fig2}, we propose a new transformer block as the basic unit to build our UDC Restoration Module. The proposed transformer block considers both global and local information for feature processing. Specifically, in the proposed transformer block, we first use a pixel-wise convolution~\cite{mobilenets} and a depth-wise convolution~\cite{efficientnet} to generate query, key, and value which are helpful to enrich the local context. Then, we perform self-attention across channels~\cite{xcit} to generate an attention map encoding the global context implicitly. Finally, we resort to the gating mechanism in the feed-forward network named Gating Feed Forward Network (GFFN) to further improve the expressive capability of networks. Specifically, the gating mechanism is formulated as the element-wise product of two parallel paths. In the first path, we adopt a pixel-wise convolution and a depth-wise convolution to strengthen the locality. In the second path, we adopt a pixel-wise convolution followed by a Sigmoid function as a gating signal. The UCD restoration module can be formulated as
\begin{equation}
F_{1}=Transformer(F_{0}),
\end{equation}
where $Transformer(\cdot)$ is the UDC Restoration module that contains several transformer blocks, $F_{0}$ represents output features of the shallow feature extractor, and $F_{1}$ is output features.

\begin{figure}[t] 
\centering
\includegraphics[width=0.42\textwidth]{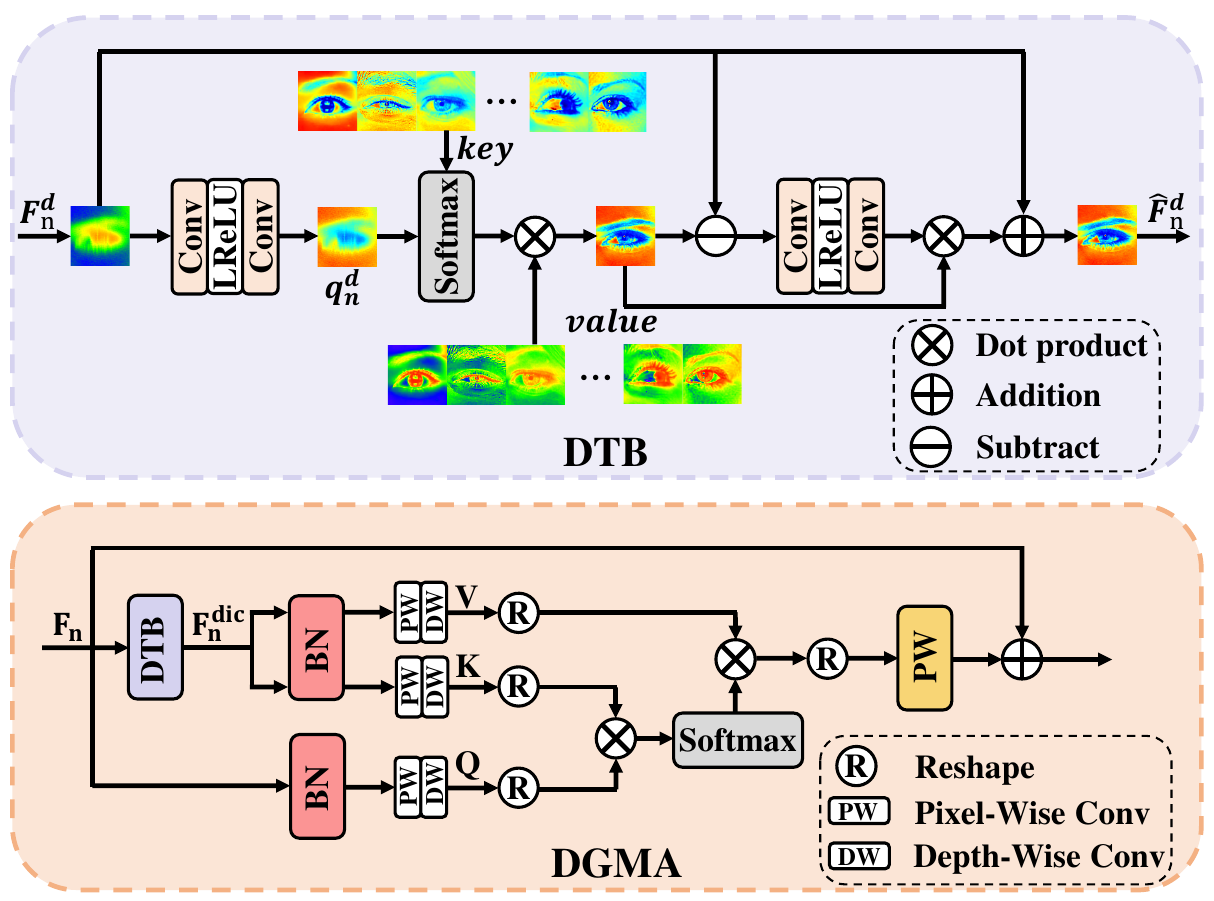}
% %\vspace{-2mm}
\caption{The structure of the proposed Dictionary Transform Block (DTB) and Dictionary-Guided Multi-head Attention (DGMA). For simplicity, we show only the process of DTB transforming the features of the right eye component. The transformation of other component features, RoIAlign, and reverse RoIAlign operation are omitted.}
\label{fig3}
%\vspace{-4mm}
\end{figure}

\begin{figure*}[t]
\centering
\includegraphics[width=0.95\textwidth]{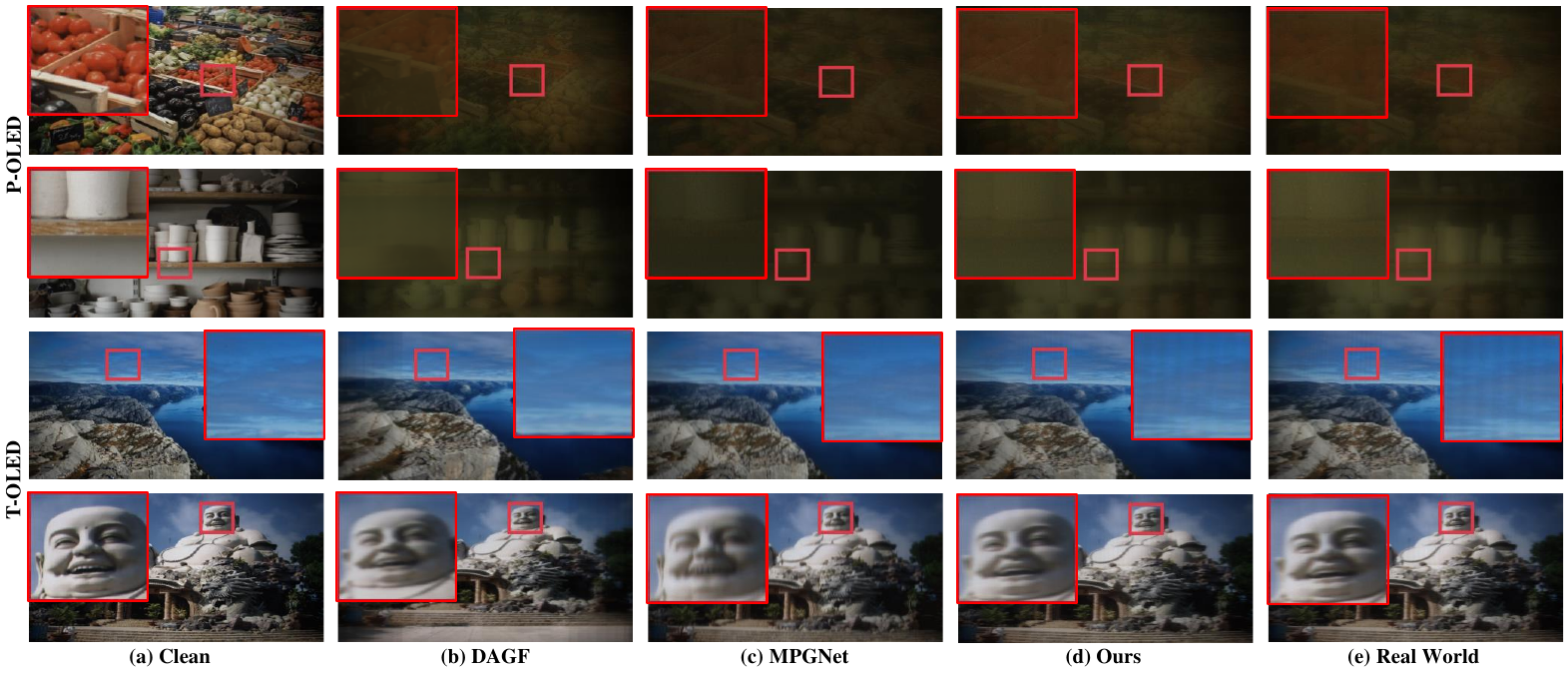}
%\vspace{-2mm}
\caption{Visual comparison of the synthetic UDC image samples. (a) clean images. (b-d) Synthetic UDC images by DAGF, MPGNet, and Ours, respectively. (e) The real-world captured UDC images.}
\label{fig4}
%\vspace{-4mm}
\end{figure*}

\subsubsection{Dictionary-guided Restoration Module}
To achieve fine-grained UDC face restoration, we exploit the face dictionary as extra facial prior knowledge in the proposed dictionary-guided restoration module to further help restore face details of the face images. 
The dictionary-guided restoration module is designed under an encoder-decoder structure with skip connections, which can fully exploit both multi-scale features and face dictionary prior for effective face restoration. 
In our dictionary-guided restoration module, we propose a novel Dictionary-Guided Transformer Block (DGTB) as the basic block of the encoder-decoder structure to restore the facial details of the image. As shown in Fig.~\ref{fig2}, the DGTB consists of a Dictionary-Guide Muti-head Attention (DGMA), a GFFN in the UDC Restoration Module, and several transformer blocks. DGMA has facial component dictionaries and queries these dictionaries through Dictionary Transform Block (DTB). By storing the high-quality textures of the eyes, nose, and mouth in the facial component dictionary in advance, we can significantly improve the model's ability to restore face details.

DGMA is our core component, which is shown in Fig.~\ref{fig4}. The facial component dictionary $Dic$ in each DGMA can be expressed as follows
\begin{equation}
    Dic={(key_1:value_1),...,(key_y:value_y)},
\end{equation}
where $y$ is the number of key values in the dictionary. To make sure the dictionary can be end-to-end optimized by the final loss, we set the $(key, value)$ in the dictionary as learnable parameters. The size of the $key$ and $value$ are related to the scale where DGMA is located. Taking the eye dictionary as an example, in the DGMA with an input feature size of $(\frac{H}{s}\times\frac{W}{s}\times sC)$, the size of each key and value is $(\frac{A^{eye}}{s} \times\frac{A^{eye}}{s})$, where $A^{eye}$ is the size of the detected eye landmark box.
Supposing the input feature of DGMA is $F_n$, DGMA first takes the obtained $F_n$ into DTB. DTB first obtains facial component features $F^{d}_{n}$ such as eyes and mouth from $F_n$ through RoIAlign operation and detected facial landmarks. Then query features $q^{d}_{n}$ is obtained through a series of convolution operations on features $F^{d}_{n}$. 
Then DTB queries the component dictionary in a similar way as the attention mechanism. After that, DTB fuses the original component features and the component features queried in the dictionary, then copies and pastes fused component features $\hat{F^{d}_{n}}$ into the original features $F_n$ through reverse RoIAlign operation to obtain features ${F_n^{dic}}$. 
Then, instead of using self-attention directly to features ${F_n^{dic}}$, our DGMA takes input features $F_n$ as queries $Q=W_pW_{d}BN(F_n)$, while the keys $K=W_pW_{d}BN(F_n^{dic})$ and values $V=W_pW_{d}BN(F_n^{dic})$ are generated from output feature of DTB $F_n^{dic}$, where $W_p$ is a pixel-wise convolution and $W_d$ is a depth-wise convolution. Finally, we adopt the attention mechanism to conduct face restoration. Overall, the DGMA process is defined as
\begin{equation}
    \hat{F_n}=W_p(softmax(KQ/\beta)V)+F_n,
\end{equation}
where $\beta$ is a learnable scaling parameter. With the guidance of the face dictionary prior, our DGMA can fuse the information from the features $F_n$ and its corresponding high-quality face texture features $F_n^{dic}$. They can respectively provide identity information and high-quality facial details for face restoration. While normal self-attention usually can only consider the feature of a single information source.

For the DGTB of each scale, after using a DGMA and a GFFN, we choose to use several transformer blocks mentioned in UDCRM to refine the features of this scale instead of continuing to use DGMA. Because we find that the use of the DGMA query dictionary repeatedly can cause model performance to decline.

\subsubsection{Image Reconstruction Module}
Following the Dictionary Guided Restoration Module, deep features $F_{d} \in\mathbb{R}^{H\times W\times C}$ are put into the image reconstruction module. In this module, we use four consecutive transformer blocks and a $3\times3$ convolution to transform $F_{d}$ into the final clear face image. To simplify our DGFormer, we use the same transformer block that is used in the UDC restoration module. 

\subsubsection{Loss Function}
To train our DGFormer, we use L1 loss between the recovered images and the ground truth images.

\begin{figure*}[t]
\centering
\includegraphics[width=0.95\textwidth]{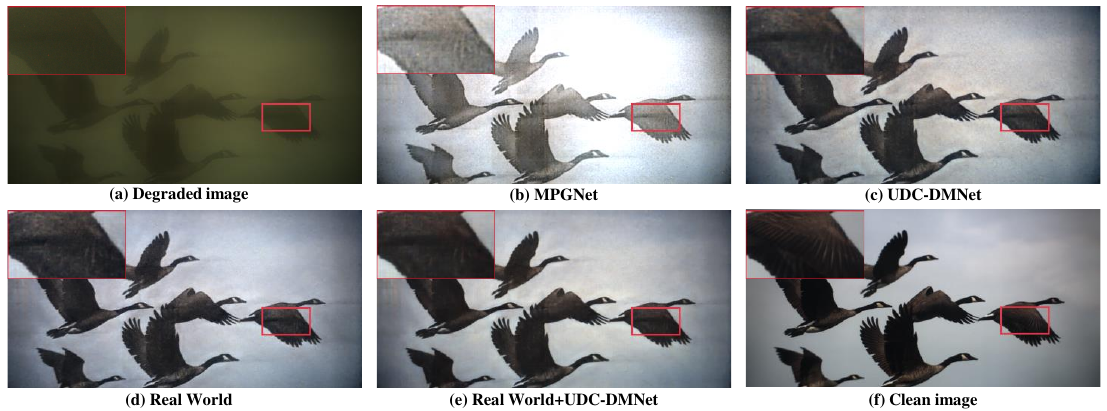}
%\vspace{-2mm}
\caption{Visual comparison of restoration results of NAFNet trained on different synthetic and real-world P-OLED datasets. (a) The real-world captured UDC images. (b-c) NAFNet trained on different synthetic P-OLED datasets. (d) NAFNet trained on real-world P-OLED datasets. (e) NAFNet trained on real-world -POLED datasets and our synthetic POLED datasets. (f) clean images.}
\label{fig5}
%\vspace{-4mm}
\end{figure*}

\begin{figure*}[t]
\centering
\includegraphics[width=0.95\textwidth]{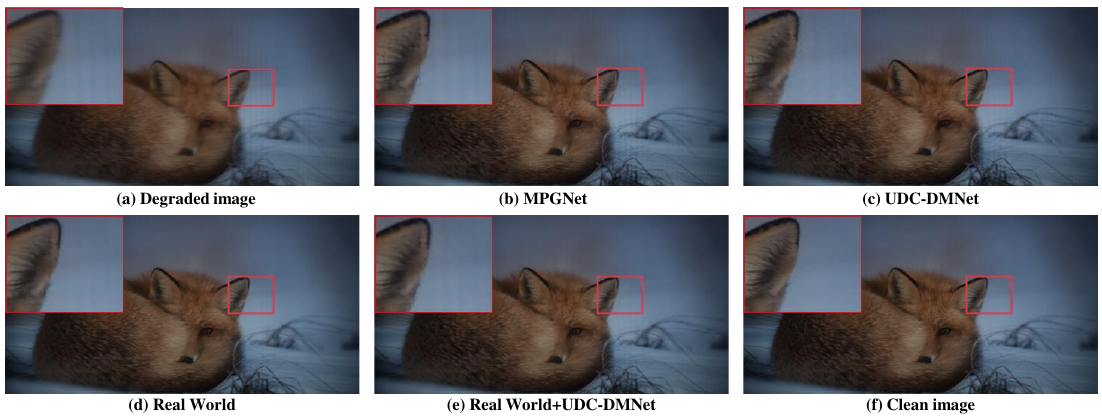}
%\vspace{-2mm}
\caption{Visual comparison of restoration results of NAFNet trained on different synthetic and real-world T-OLED datasets. (a) The real-world captured UDC images. (b-c) NAFNet trained on different synthetic T-OLED datasets. (d) NAFNet trained on real-world T-OLED datasets. (e) NAFNet trained on real-world T-OLED datasets and our synthetic T-OLED datasets. (f) clean images}
\label{fig6}
%\vspace{-4mm}
\end{figure*}

\section{Experiment}
\label{sec:experiment}
In this section, we conducted a series of experiments to compare our synthesis methods with the existing synthesis methods and the real-world datasets, and compare our restoration methods with existing UDC restoration and face restoration methods. The results of quantitative and qualitative experiments prove the effectiveness of both synthesis and restoration methods. Furthermore, we have performed an ablation study to validate the efficacy of each module within our network architecture.

\subsection{Implementation Details}
In this work, the training settings of the two networks are different. For UDC-DMNet, we train it using Adam optimizer~\cite{adam} with $\beta_1=0.9$, $\beta_2=0.99$ to minimize a weighted combination of PSNR loss~\cite{hinet}, perceptual loss~\cite{perceptual}, and the adversarial loss~\cite{geometric} for $8\times10^4$ iterations. We choose PatchGANDiscriminator~\cite{cyclegan} as the discriminator. The initial learning rates are set to both $1\times10^{-4}$ for the generator and discriminator. The learning rate is decayed with a cosine annealing schedule~\cite{sgdr}, where the learning rate is decreased to $1\times10^{-7}$. The batch size is set to $8$, and we apply horizontal flipping and rotation as data augmentation and crop image patch of $512\times512$ for training. And as for DGFormer, the number of transformer blocks in the UDC Restoration Module and Image Reconstruction Module are both $4$. The number of transformer blocks in each layer of the Dictionary Guided Restoration Module is $\{6, 6, 8, 6, 6\}$, and attention heads are $\{2, 4, 8, 4, 2\}$. We use AdamW~\cite{decoupled} optimizer ($\beta_1=0.9$, $\beta_2=0.999$, weight decay 0.01) with the cosine annealing strategy to train the DGFormer, where the learning rate gradually decreases from the initial learning rate $2\times10^{-4}$ to $2\times10^{-6}$ for $40$ epochs. We choose $L_1$ loss and perceptual loss as the final loss and the training batch size is set to $10$.

\begin{table}[t]
    \caption{Quantitative comparisons among different UDC degradation synthesis methods. We train the representative restoration method NAFNet~\cite{nafnet} on the synthesized P-OLED and T-OLED data from different synthesis methods and report its restoration performance on corresponding testing sets.
    Data synthesized by UDC-DMNet demonstrate the optimal utility with regard to real-world data.}
    %\vspace{-1mm}
    \centering
    \begin{tabular}{lcccc}
    \hline
        \multirow{2}{*}{Data from} & \multicolumn{2}{c}{P-OLED~\cite{ECCVchallenge}}  & \multicolumn{2}{c}{T-OLED~\cite{ECCVchallenge}}  \\ 
        & PSNR$\uparrow$ & SSIM$\uparrow$ & PSNR$\uparrow$ & SSIM$\uparrow$ \\ \hline
        Real world~\cite{ECCVchallenge} & 27.79 & 0.8446 & 32.98 & 0.9248 \\ 
        DAGF~\cite{DAGF} & 21.20 & 0.6554 & 24.20 & 0.8031 \\ 
        MPGNet~\cite{MPGNet} & 21.77 & 0.7471 & 32.66 & 0.9240 \\ 
        UDC-DMNet & 27.03 & 0.8217 & 32.86 & 0.9244 \\ 
        Real world+UDC-DMNet & 28.36 & 0.8595 & 33.93 & 0.9375 \\ \hline
    \end{tabular}
    \label{generate}
    \vspace{-4mm}
\end{table}

\begin{figure*}[t]
\centering
\includegraphics[width=0.92\textwidth]{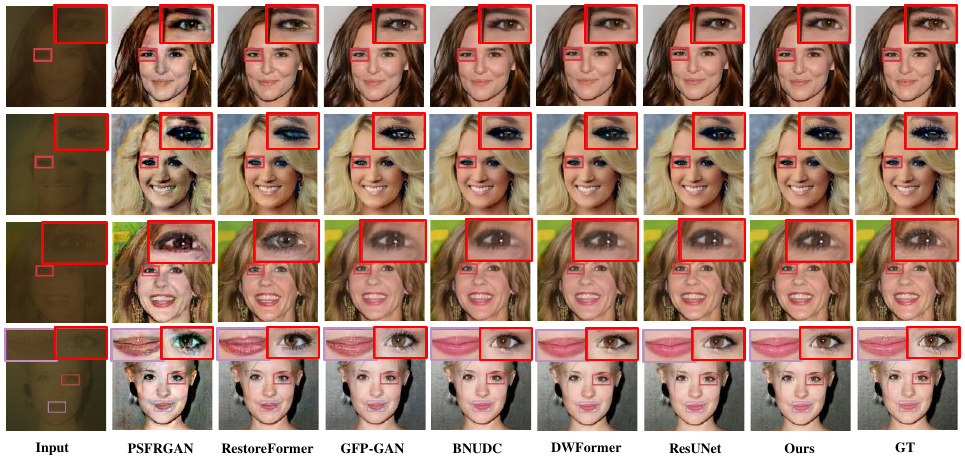}
%\vspace{-2mm}
\caption{Visual comparison results of CelebA-Test-P. We compare face restoration methods: PSFRGAN~\cite{PSFRGAN}, RestoreFormer~\cite{Restoreformer}, GFP-GAN~\cite{GFPGAN} and UDC restoration methods: BNUDC~\cite{BNUDC}, DWFormer~\cite{MPGNet}, ResUNet~\cite{ResNet}. Our method produces more faithful details.}
 \label{fig7}
 %\vspace{-4mm}
\end{figure*}

\begin{figure*}[t]
\centering
\includegraphics[width=0.92\textwidth]{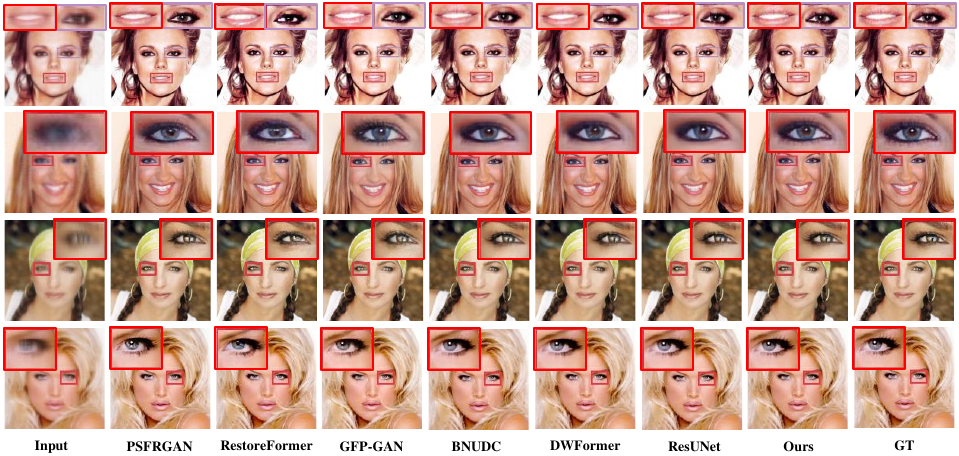}
%\vspace{-4mm}
\caption{Visual comparison results of CelebA-Test-T. We compare face restoration methods: PSFRGAN~\cite{PSFRGAN}, RestoreFormer~\cite{Restoreformer}, GFP-GAN~\cite{GFPGAN} and UDC restoration methods: BNUDC~\cite{BNUDC}, DWFormer~\cite{MPGNet}, ResUNet~\cite{ResNet}. Our method produces more faithful details.}
 \label{fig8}
 %\vspace{-4mm}
\end{figure*}

\subsection{Analysis of Synthesis Method}
Since our UDC-DMNet is proposed to simulate a degradation process in generating UDC images, in this work, we mainly compare with two related works, \ie DAGF~\cite{DAGF}, MPGNet~\cite{MPGNet} which are both representative methods for synthesizing UDC images. Besides, we also compare with real-world UDC images~\cite{ECCVchallenge}.

The qualitative comparisons are shown in Fig.~\ref{fig4}. For the P-OLED scene, the results demonstrate that DAGF cannot simulate the noise, and P-OLED UDC images synthesized by DAGF are over-smooth. In this regard, images generated by our method and MPGNet are closer to the ground truth. While MPGNet cannot simulate the brightness of P-OLED UDC images. As the results show, the P-OLED UDC images synthesized by MPGNet tend to be darker. Besides, MPGNet and DAGF cannot simulate color-shift well. For the T-OLED scene, the blurred details of the UDC images synthesized by MPGNet and DAGF are different from the real-world UDC images. And they cannot simulate the striped shadows in T-OLED UDC images compared with ours.

We use different synthetic datasets and real-world UDC datasets to train an identical restoration model, NAFNet~\cite{nafnet}. For a fair comparison, the model is trained on different data sources by $200$ epochs with the same training parameters. After training, we evaluate the restoration model on real-world UDC images. 
The quantitative results for different data sources are shown in Table~\ref{generate}. For P-OLED, the model trained with our synthetic dataset achieves the closest PSNR and SSIM to the real-world dataset and is significantly higher than other results with synthetic datasets generated by DAGF and MPGNet. For T-OLED, the model trained with our dataset still slightly outperforms others and is closest to the real-world dataset. At the same time, when we combine our dataset and real-world dataset to train the restoration model, the performance of the model is improved compared to the counterpart with only the real-world dataset. It shows that the dataset we build can play the role of supplement and thus improve the generalization of the model. 
The qualitative results across diverse data sources are shown in Fig~\ref{fig5} and~\ref{fig6}. For P-OLED, when compared to MPGNet, images restored by NAFNet, trained with UDC-DMNet's synthetic P-OLED datasets, perform a closer match in brightness to images restored by NAFNet trained with real-world datasets. Furthermore, images restored by NAFNet, trained with both synthetic and real-world datasets, are closer to clean images in comparison to NAFNet trained only on real-world datasets.
In the case of T-OLED, NAFNet trained with UDC-DMNet's synthetic T-OLED datasets performs superior capability in removing striped shadows when contrasted with MPGNet. Simultaneously, the synthetic dataset serves to complement the real-world datasets, thereby contributing to the overall enhancement of the restoration model's performance.

% The qualitative results for each dataset are shown in Fig. 6, 

\subsection{Comparison with SOTA Restoration Methods}
\begin{table}[t]
    \caption{Quantitative comparison on synthetic CelebA-Test-P for blind face restoration. Bold font and underline indicate the best and the second-best performance.}
    %\vspace{-1mm}
    \centering
    \begin{tabular}{lccccc}
    \hline
        Method & PSNR$\uparrow$ & SSIM$\uparrow$ & LPIPS$\downarrow$ & Deg.$\downarrow$ & LMD$\downarrow$ \\ \hline
        input & 9.84 & 0.4448 & 0.6808 & 68.89 & 3.50 \\ 
        GFPGAN\cite{GFPGAN} & 28.04 & 0.8359 & 0.2766 & 22.71 & 1.17 \\ 
        VQFR\cite{VQFR} & 24.14 & 0.6629 & 0.3270 & 43.40 & 2.72 \\ 
        PSFR-GAN\cite{PSFRGAN} & 23.39 & 0.7001 & 0.4464  & 37.36 & 1.87 \\ 
        RestoreFormer\cite{Restoreformer} & 24.66 & 0.6994 & 0.3349 & 27.91 & 1.76 \\ \hline
        BNUDC\cite{BNUDC} & 31.58  & 0.8993 & 0.2342 & 21.03 & 0.96 \\ 
        DAGF\cite{DAGF} & 30.47 & 0.8678 & 0.2909  & 22.41 & 1.07 \\ 
        DWFormer\cite{MPGNet} & 30.83 & 0.8869 & 0.2629 & 22.96 & 1.05 \\ 
        ResUNet\cite{ResNet} & \underline{31.84} & \underline{0.9010} & \underline{0.2280} & \underline{20.66} & \underline{0.92} \\ \hline
        DGFormer (Ours) & \textbf{32.69} & \textbf{0.9094} & \textbf{0.2015} & \textbf{18.97} & \textbf{0.85} \\ \hline
    \end{tabular}
    \label{FFHQ-P}
    %\vspace{-2mm}
\end{table}

We compare the performance of the proposed DGFormer with several state-of-the-art UDC restoration methods: BNUDC~\cite{BNUDC}, DAGF~\cite{DAGF}, ResUNet~\cite{ResNet}, DWFromer~\cite{MPGNet} and several state-of-the-art face restoration methods: GFPGAN~\cite{GFPGAN}, PSFRGAN~\cite{PSFRGAN}, VQFR~\cite{VQFR}, RestoreFormer~\cite{Restoreformer} on the synthetic UDC face dataset. 
To ensure a fair comparison, all methods are trained under the same training settings. 
For evaluation, we employ pixel-wise metrics (PSNR and SSIM) and the perceptual metric (LPIPS~\cite{lpips}). And we measure the identity distance with angels in the ArcFace~\cite{arcface} feature embedding as the identity metric, which is denoted by ‘Deg.’. We also adopt landmark distance (LMD) as the fidelity metric to better measure the fidelity with accurate facial positions and expressions.

The quantitative comparisons in terms of the above metrics are reported in Table~\ref{FFHQ-P} and Table~\ref{FFHQ-T}. The results show that our method achieves state-of-the-art performance on both CelebA-Test-P and CelebA-Test-T. Specifically, DGFormer achieves the best performance regarding PSNR and SSIM. Moreover, DGFormer achieves the lowest LPIPS, indicating that the perceptual quality of restored faces is closest to ground truth. DGFormer also achieves the best Deg. and LDM, showing that it can recover accurate facial parts and details. 
In addition, some restoration methods based on face priors perform poorly. 
We suspect that this is mainly due to the difference between UDC degradation and normal degradation. UDC degradation is a combination of multiple degradations such as low light, blur, and noise, which is distinct from normal degradation. 
Therefore, if the specific characteristics of the UDC image are not considered when introducing the face prior in the network, the prior may not work well.

\begin{table}[t]
    \caption{Quantitative comparison on synthetic CelebA-Test-T for blind face restoration. Bold font and underline indicate the best and the second-best performance.}
    %\vspace{-1mm}
    \centering
    \begin{tabular}{lccccc}
    \hline
        Method & PSNR$\uparrow$ & SSIM$\uparrow$ & LPIPS$\downarrow$ & Deg.$\downarrow$ & LMD$\downarrow$ \\ \hline
        input & 25.56 & 0.7673 & 0.3796 & 32.22 & 1.99 \\ 
        GFPGAN\cite{GFPGAN} & 31.09 & 0.9479 & 0.1705 & 13.01 & 0.78 \\ 
        VQFR\cite{VQFR} & 24.84 & 0.6841 & 0.3354 & 31.67 & 2.04 \\ 
        PSFR-GAN\cite{PSFRGAN} & 31.09 & 0.8750 & 0.2294  & 15.43 & 0.86 \\ 
        RestoreFormer\cite{Restoreformer} & 27.42 & 0.7505 & 0.2419 & 16.22 & 1.00 \\ \hline
        BNUDC\cite{BNUDC} & 36.49  & 0.9498 & 0.1232 & 8.98 & 0.36 \\ 
        DAGF\cite{DAGF} & 33.63 & 0.9131 & 0.1826  & 11.79 & 0.48 \\ 
        DWFormer\cite{MPGNet} & 36.68 & 0.9533 & 0.1194 & 8.72 & 0.35 \\ 
        ResUNet\cite{ResNet} & \underline{37.38} & \underline{0.9604} & \underline{0.0951} & \underline{7.55} & \underline{0.30} \\ \hline
        DGFormer (Ours) & \textbf{38.35} & \textbf{0.9678} & \textbf{0.0720} & \textbf{6.49} & \textbf{0.27} \\ \hline
    \end{tabular}
    \label{FFHQ-T}
    %\vspace{-4mm}
\end{table}

The qualitative results are presented in Fig.~\ref{fig7},~\ref{fig8}. Compared with UDC restoration methods, thanks to our special design dictionary-guided restoration module, our method recovers faithful details in the eyes, mouth, \etc On the contrary, face images restored by normal UDC restoration methods are over-smooth and lose facial details. 
Compared with face restoration methods, some of them do not remove the degradation of UDC face images well (see the second column in Fig.~\ref{fig7},~\ref{fig8}). And some of them introduce texture information that does not belong to the original face (see the third and fourth columns in Fig.~\ref{fig7},~\ref{fig8}). In contrast, our method performs better in both aspects.
This is attributed to the fact that the characteristics of UDC degradation are taken into consideration when designing the network.

\begin{figure*}[t]
\centering
\includegraphics[width=0.92\textwidth]{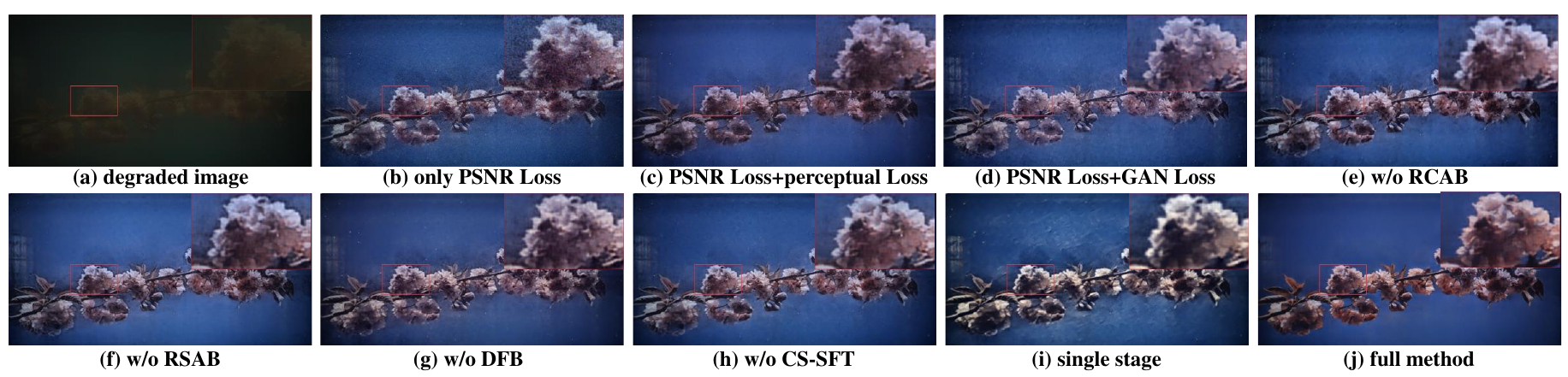}
%\vspace{-4mm}
\caption{Visual comparison results of ablation studies on UDC-DMNet.  (a) The real-world captured UDC image. (b-d) NAFNet trained on different P-OLED datasets synthesized by UDC-DMNet with different losses. (e-i) NAFNet trained on different P-OLED datasets synthesized by UDC-DMNet with different configurations. (j) NAFNet trained on P-OLED datasets synthesized by UDC-DMNet.}
\label{fig9}
%\vspace{-4mm}
\end{figure*}

\begin{figure}[t]
\centering
\includegraphics[width=0.45\textwidth]{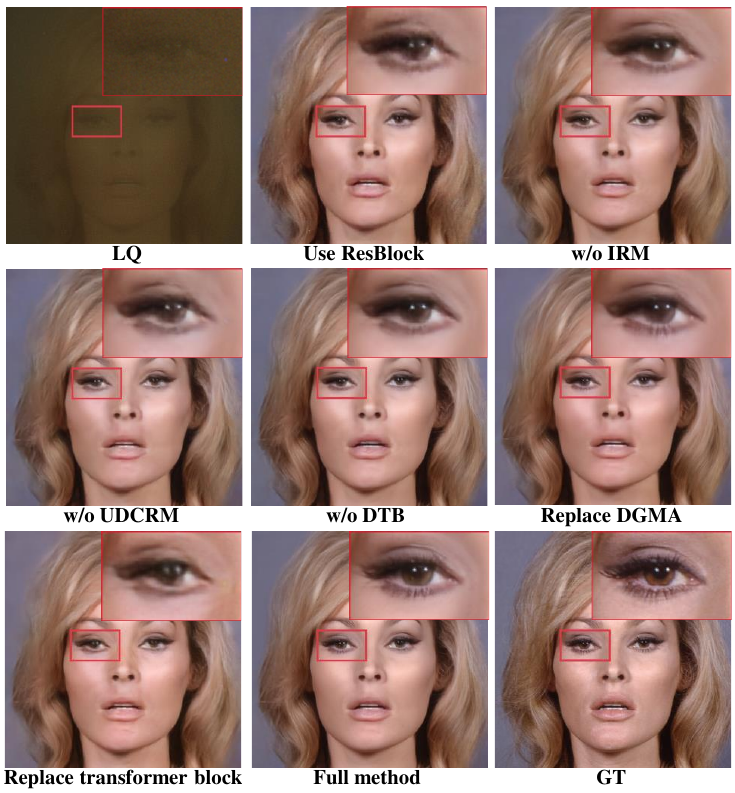}
%\vspace{-4mm}
\captionsetup{justification=centering}
\caption{Visual comparison results of ablation studies on DGFormer. }
\label{fig10}
\vspace{-4mm}
\end{figure}

\subsection{Ablation Studies}
\begin{table}[t]
    \caption{Ablation studies of UDC-DMNet.}
    \centering
    %\vspace{-1mm}
    \begin{tabular}{lcc}
    \hline
        Configurations & PSNR$\uparrow$ & SSIM$\uparrow$ \\ \hline
        only PSNR Loss & 24.31 & 0.6179 \\
        PSNR Loss + Perceptual Loss & 26.10 & 0.7917 \\ 
        PSNR Loss + GAN Loss & 26.69 & 0.8131 \\ 
        \hline
        w/o RSAB & 26.02 & 0.8052 \\ 
        w/o RCAB & 25.92 & 0.7989 \\ 
        w/o CSSFT & 26.58 & 0.7943 \\
        w/o DFB & 25.85 & 0.7896 \\
        Single Stage & 24.96 & 0.7885 \\ 
        \hline
        Full method & 27.03 & 0.8217 \\ \hline  
    \end{tabular}
    \label{UDC-DM}
    %\vspace{-5mm}
\end{table}
\subsubsection{UDC-DMNet}
In the ablation of UDC-DMNet, we analyze the effectiveness of its components. We use UDC-DMNet with different settings to synthesize different P-OLED datasets to train an identical restoration model, NAFNet~\cite{nafnet}. For a fair comparison, the model is trained on different datasets by $200$ epochs with the same training parameters. After training, we evaluate the restoration model on real-world P-OLED UDC images. Table~\ref{UDC-DM} reports the performance of different settings. 

We first perform ablation experiments on the loss function. When using only PSNR loss to train the network, the performance of the network declines significantly. The reason is that images generated by UDC-DMNet are too smooth, and UDC-DMNet fails to generate any noise degradation in the generated images. The restoration model trained with such synthetic data cannot remove UDC degradation well. Then ablation experiments are carried out on perceptual loss and adversarial loss respectively. Compared to using only PSNR loss, adding both losses can improve the utility of the synthetic dataset and make the restoration model trained by it perform better.

Secondly, we perform ablation experiments on each component in the UDC-DMNet structure. Specifically, we first remove RSAB and RCAB respectively, adopting a normal residual block~\cite{residual} in each stage. Removing RSAB causes a drop of $0.99$ dB in terms of PSNR and removing RCAB causes a drop of $1.09$ dB in terms of PSNR. It demonstrates that both kinds of attention blocks are useful for simulating degradation. Then we remove the Cross-Stage Spatial Feature Transform (CSSFT), and the performance drops by about $0.45$ dB regarding PSNR. Then we remove the Degradation Fusion Block (DFB), and the performance drops by about $1.16$ dB in terms of PSNR. All the ablation studies demonstrate the effectiveness of our proposed UDC-DMNet for UDC image synthesis. Furthermore, we modify the network to single-stage to explore the effectiveness of multi-stage modeling of the degradation process. This leads to a significant drop in performance, indicating that multi-stage modeling affects the synthesis performance significantly.
The qualitative results for different configurations are shown in Fig.~\ref{fig9}. UDC-DMNET, without any of the concerned components, can hardly simulate real-world UDC degradation. As shown in the figure, this leads to poor performance of the restoration model trained on these synthetic datasets.

\begin{table}[t]
    \caption{Ablation studies of our DGFormer.}
    \centering
    %\vspace{-1mm}
    \begin{tabular}{lccccc}
    \hline
        Configuration & PSNR$\uparrow$ & SSIM$\uparrow$ \\ \hline
        use ResBlock & 30.88 & 0.8870 \\ 
        w/o UDCRM & 32.37 & 0.9062 \\
        w/o IRM & 32.26 & 0.8998 \\
        w/o DTB in DGTB & 32.20 & 0.9038 \\  
        Replace DGMA with self-attention & 32.11 & 0.9012 \\ 
        Replace transformer blocks with DGMA+GFFN & 31.89 & 0.9002 \\  
        Full method & 32.69 & 0.9094 \\ \hline
    \end{tabular}
    \label{DGFormer}
    %\vspace{-4mm}
\end{table}
\subsubsection{DGFormer}
We further analyze and discuss the effectiveness of the internal modules in our DGFormer. Table~\ref{DGFormer} shows the results, where all models are trained on FFHQ-P and tested on CelebA-Test-P.

Firstly, we replace the transformer blocks with residual blocks~\cite{resblock,residual} for training, leading to a significant drop in performance. It implies that transformer blocks are important for the integration of global information and local information. Secondly, we remove the UDC Restoration Module (UDCRM) and Image Reconstruction Module (IRM) respectively, which both result in a performance drop, implying that both coarse-grained UDC degradation removal and image feature refinement are indispensable.
Thirdly, we remove the Dictionary Transform Block (DTB) in each Dictionary-Guided Transformer Block (DGTB). In this way, we will not use the facial component dictionary in the network.
This also causes performance drop, and the facial details are easily lost, demonstrating that dictionaries are helpful in the restoration of facial component details.
Fourth, we replace DGMA with normal self-attention, which results in a performance drop. This shows that DGMA, which combines low-quality features and features enhanced by prior information, is more conducive to image restoration than self-attention using only a single feature as an information source.
Finally, we replace the normal transformer block in DGTB with DGMA + GFFN, causing a performance drop, which shows that the repeated use of the dictionary has a negative impact on the restoration of details.

The visual results of varying configurations are presented in Fig.~\ref{fig10}. Regarding `use Resblocks' and `w/o UDCRM' in the figure, the absence of a transformer structure and UDCRM cannot fully eliminate UDC degradation, leading to blurring and noise artifacts in the restored images. 
For `w/o DTB' in the figure, the restoration model without the dictionary component struggles to effectively restore eye details.  
In addition, as shown in the `Replace transformer block' of the figure, when we replace the transformer block in DGTB with DGMA + GFFN, the eyes of the restored image become blurrier.
Furthermore, in comparison with `w/o IRM' and `Replace DGMA', images restored by full DGFormer showcase more comprehensive and detailed eye features. This illustrates the refinement effect of IRM and DGMA on image features.

\section{Conclusion and Future Work}
\label{sec:conclusion}
In this work, we tackle the UDC face restoration problem for the first time. We first introduce a two-stage network called UDC-DMNet for synthesizing UDC images by simulating the color filtering and brightness attenuation process and the diffraction process in UDC imaging. Experiments show that our UDC-DMNet can synthesize realistic UDC images, which are also superior to images synthesized by other methods in training the restoration model. 
Furthermore, we propose a novel dictionary-guide transformer named DGFormer which takes both UDC characteristics and face priors into account. 
Extensive experiments show that our DGFormer achieves superior performance than existing UDC restoration methods and face restoration methods in UDC scenarios.

In the future, we will consider the following challenges for UDC Face restoration. Firstly, considering that the predominant utilization of UDC occurs in mobile phones, it is necessary to devise a more lightweight face restoration model and optimize it for the hardware architecture of mobile devices. Secondly, it's essential to consider the evolving landscape of real-world UDC face images, where additional complexities such as glare and reflections can emerge based on the specific capture contexts. Consequently, It is still necessary to collect real-world UDC images of more scenes.

% \newpage
\bibliographystyle{IEEEtran}
\bibliography{myreference}

\end{document}